\documentclass[10pt,twocolumn,letterpaper]{article}
\usepackage{cvpr}

\usepackage{graphicx}
\usepackage{amsmath}
\usepackage{amssymb}
\usepackage{booktabs}
\usepackage{multirow}
\usepackage{multicol}
\usepackage[pagebackref,breaklinks,colorlinks,bookmarks=false]{hyperref}

\usepackage[capitalize]{cleveref}
\crefname{section}{Sec.}{Secs.}
\Crefname{section}{Section}{Sections}
\Crefname{table}{Table}{Tables}
\crefname{table}{Tab.}{Tabs.}

\usepackage{amssymb}
\usepackage{amsthm}
\usepackage{hyperref}
\hypersetup{
	colorlinks=true,
	linkcolor=cyan,
	filecolor=blue,      
	urlcolor=red,
	citecolor=green,
}
\usepackage{booktabs}
\usepackage{lineno}

\usepackage[figuresright]{rotating}
\usepackage{threeparttable}
\usepackage{cleveref}

\begin{document}



\title{Learnable Ophthalmology SAM}
\author{
   Zhongxi Qiu$^{*}$ \and Yan Hu$^{\dagger}$ \thanks{Equal contributions} \and Heng Li \and Jiang Liu \thanks{Corresponding to Yan Hu and Jiang Liu} 
   \and 
   {\tt\small \{huy3,liuj\}@sustech.edu.cn} \\
   Research Institute of Trustworthy Autonomous Systems and \\ Department of Computer Science and Engineering, \\ Southern University of Science and Technology, Shenzhen 518055, China 
}
\maketitle





\begin{abstract}
Segmentation is vital for ophthalmology image analysis. But its various modal images hinder most of the existing segmentation algorithms applications, as they rely on training based on a large number of labels or hold weak generalization ability. Based on Segment Anything (SAM), we propose a simple but effective learnable prompt layer suitable for multiple target segmentation in ophthalmology multi-modal images, named Learnable  Ophthalmology Segment Anything (SAM). The learnable prompt layer learns medical prior knowledge from each transformer layer. During training, we only train the prompt layer and task head based on a one-shot mechanism. We demonstrate the effectiveness of our thought based on four medical segmentation tasks based on nine publicly available datasets. Moreover, we only provide a new improvement thought for applying the existing fundamental CV models in the medical field. Our codes are available at \href{https://github.com/Qsingle/LearnablePromptSAM}{website}.
\end{abstract}



\section{Introduction}
\label{Introduction}
Segmentation is vital for ophthalmology diagnosis and treatment. The Department of Ophthalmology holds more than 10 kinds of imaging \cite{li2018advances}. The difference between multi-modal images brings different segmentation targets, for example, blood vessels from the color fundus, and retinal layers from optical coherence tomography (OCT), which hinders the application of a single model in ophthalmology, since most of the existing segmentation algorithms rely on the labels from experts or hold weak generalization ability.

\begin{figure*}
    \centering
    \includegraphics[width=0.9\textwidth]{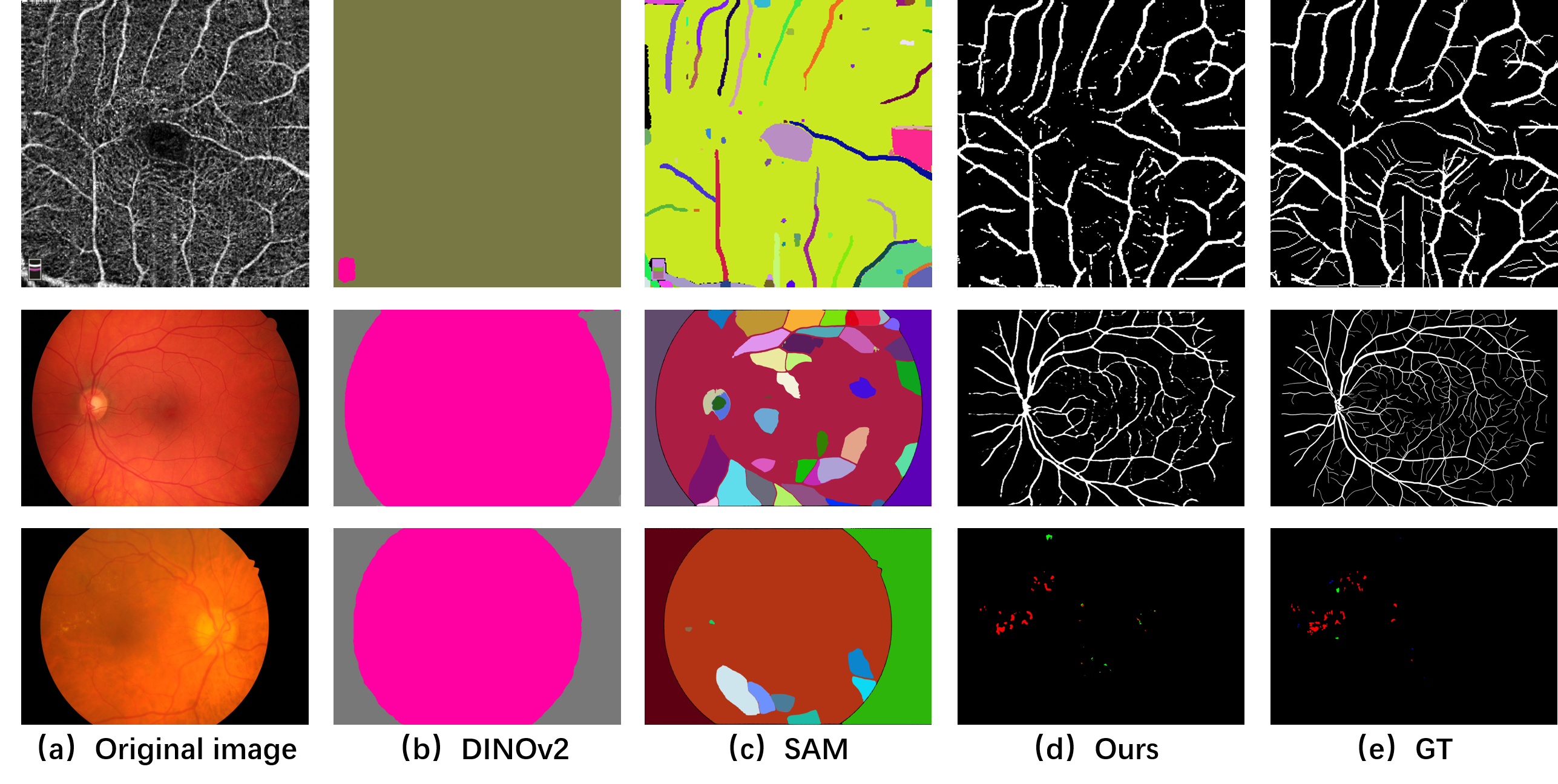}
    \caption{Examples of blood vessels and lesions extraction by fundamental CV models and our algorithm.}
    \label{fig:intro}
\end{figure*}

Some fundamental CV models of Segment Anything (SAM) \cite{kirillov2023segment}, DINOv2 \cite{oquab2023dinov2}, are released this month, which are large ViT-based model trained on the large visual corpus. Both of them have proved promising segmentation capabilities in various natural scenarios. But they cannot segmentation the blood vessel or lesions from medical images, which are helpful for doctors' diagnosis or treatment plan. As shown in Fig. \ref{fig:intro}, DINOv2 cannot provide blood vessels from the retinal color fundus or optical coherence tomography angiography (OCTA). SAM can find several blood vessels from OCTA image, but it cannot segment vessels or lesions from color fundus. The possible reason is that the edge differences between vessels or lesions and retinal are indistinct. For the retinal Optical Coherence Tomography (OCT) images, SAM neither produce layer information, which considers as a whole. Thus, SAM cannot directly apply into our medical image segmentation.

Since these fundamental CV models hole the great potential of the foundation models for computer vision, we believe they are also helpful for the medical field \cite{mazurowski2023segment, he2023accuracy}. However, their mask segmentation predictions have not been fully explored in the medical field. We analyze lots of segmentation results and find that SAM and DINOv2 provide acceptable segmentation results under the condition of distinct edge differences. However, blood vessels or lesions may be not distinctive enough to be recognized. As we know, the feature extraction abilities of the above two models are confirmed by some other computer vision tasks \cite{yu2023inpaint, ShilongLiu2023GroundingDM}. What is the best way to adapt such a foundation model to medical image segmentation in terms of effectiveness and efficiency?

Model fine-tuning is often applied to use such foundation models on large-scale benchmarks. Full-tuning the parameters in the entire network or head-tuning by only optimizing the model's head cannot provide available results in the medical field. Recently, prompt tuning has achieved considerable results in Natural Language Processing \cite{liu2023pre} and natural image processing \cite{liu2022prompt}, and its definition is treating the prompts as task-specific continuous vectors and directly optimizing them via gradients during fine-tuning \cite{liu2021p}. For example, two sets of prompts used in SAM, including sparse (points, boxes, text) and dense (masks) cannot provide accurate segmentation results in the medical field, especially for blood vessels. Therefore, in the paper, we propose a new learnable prompt for SAM, which accurately specifies what to segment in a medical image. 

Therefore, the contributions of the paper are listed as:

1. We propose a new learnable prompt layer for SAM, named Learnable Ophthalmology SAM, which accurately segments blood vessels or lesions or retinal layers in multi-modal ophthalmology images after one-shot fine-tuning.

2. The proposed learnable prompt is able to automatically learn its interested target in different modal images and hold generalization ability between datasets.

3. We demonstrate the effectiveness of the proposed prompt on four segmentation tasks based on nine publicly available datasets.

\section{Learnable Ophthalmology SAM}
\label{Method}
Inspired by \cite{10.1007/978-3-031-19827-4_41, liu2022prompt}, we freeze the main parameter of the backbone model like Vision Transformer (ViT) and try to insert some task-specific learnable parameters during training, which makes the model be applicable for the downstream-tasks without full fine-tuning the model. Therefore, in this paper, we propose a simple but effective way to learn the prompt from the features, which is applied in medical image segmentation. 

\subsection{Preliminaries}
\label{Preliminaries}
For a visual transformer(ViT), an input $x$ is first processed by patch embedding, which extracts embedding features. Then $N$ transformer layers encode the extracted features to one feature representation. Finally, a task head, such as segmentation head, generates the task-specific output based on the above representation. We formulate the above process as:
\begin{equation}
    \hat{y} = H_{t}(T_N(T_{N-1}(...T_1(P(x)))))
\end{equation}
where $\hat{y}$ is the task-specific output, $H_t$ is the head for the task $t$, $T$ is the transformer block, $P$ is the patch embedding layer. 

\subsection{Learnable Prompt Layer}
\begin{figure}
    \centering
    \includegraphics[width=0.5\textwidth]{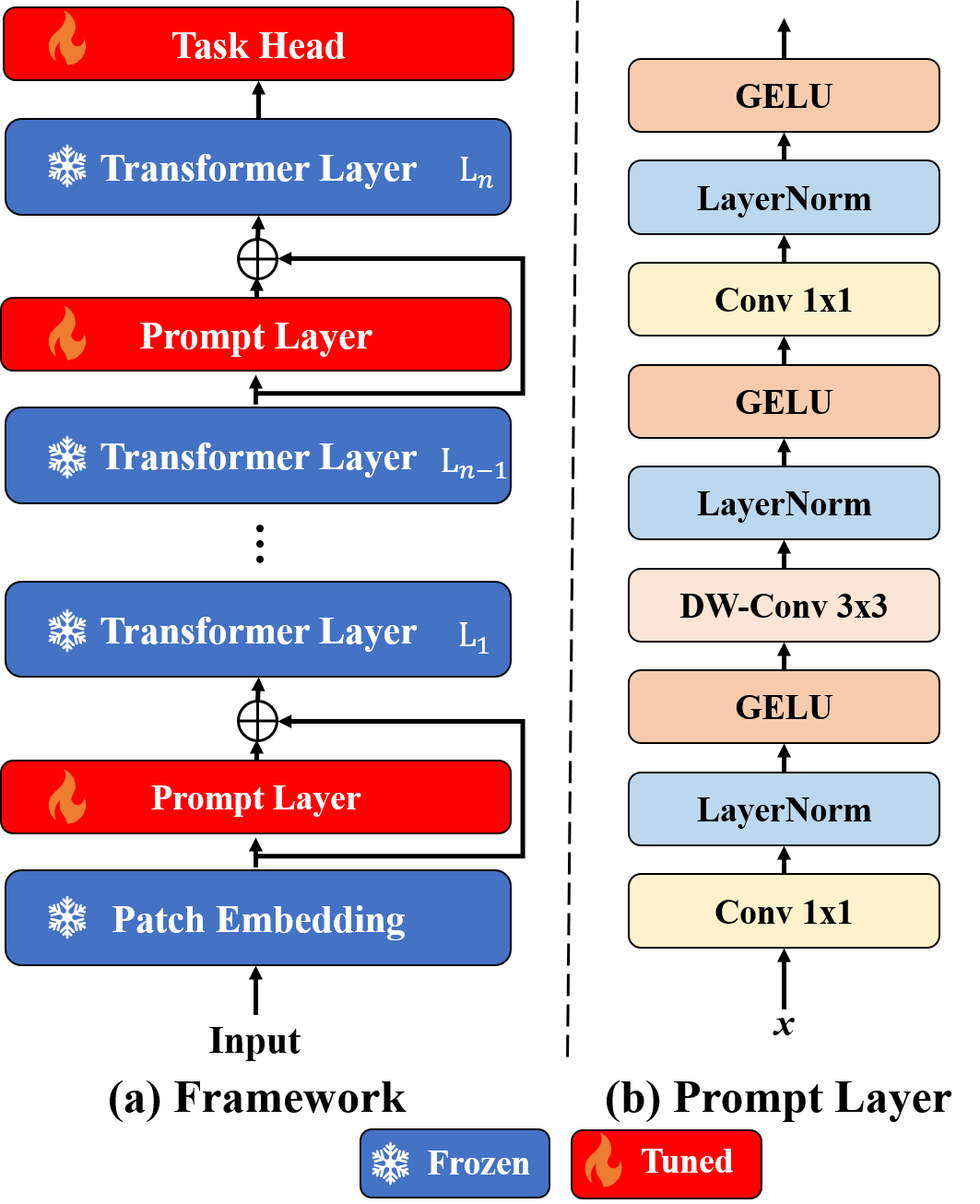}
    \caption{Structure of the framework and our proposed prompt layer.}
    \label{fig:structure}
\end{figure}

As Fig. \ref{fig:structure} (a) shows, we freeze the weights of all the transformer layers of the image encoder in the SAM. Then a task-specific head instead of the prompt generator and mask decoder of SAM is adapted to produce results. SAM is trained mainly based on natural images. Although several researchers show its certain segmentation ability in medical images \cite{mazurowski2023segment, he2023accuracy}, it cannot produce available results for some medical images, such as OCT, and color fundus. It is assumed that SAM lacks prior knowledge of these scenarios. In this paper, we propose a learnable prompt layer between each transformer layer to learn the knowledge, which is adopted as the task-specific prompt. During training, we only train the prompt layer and the task-specific head, colored with red and marked with fire, as shown in Fig. \ref{fig:structure} (a). The transformer layers and patch embedding are frozen colored blue and marked with snowflakes.

The structure of our learnable prompt layer is shown in Fig. \ref{fig:structure}(b), which is simple but effective. We just adopt two $1\times1$ Conv (Conv 1x1) with the layer normalization (LayerNorm) and GELU as the nonlinear activation function as the main part of the layer. Inspired by \cite{li2022rethinking}, $3\times3$ depthwise convolution (DW-Conv 3x3) is introduced to capture the local pattern of the features. Assumed $f_i$ ad the input features of $i-th$ transformer layer, the $i-th$ prompt layer is formulated as:
\begin{equation}
    Prompt_i = \delta(LN(W_1(\delta(LN(DW_3(\delta(LN(W_1(f_i)))))))))
\end{equation}
where $\delta$ is the GELU, $W_1$ is the weights of the $1\times1$ Conv, $DW_3$ is the weights of depthwise $3\times3$ Conv.

For the task head, we first upsample the features of the last layer in ViT twice using two transpose convolutional layers with a kernel size of $2\times 2$. Then we adopt a multi-scale convolutional layer with $group=4$ \cite{9995545} and linear layer to generate the segmentation results.




\section{Experiment Results}
\label{Experiments}

\subsection{Datasets}
We evaluate our algorithm on three medical segmentation tasks, including blood vessel segmentation, lesion segmentation, layering based on nine publicly available datasets, including three color fundus datasets FIVES \cite{Jin2022}, HRF \cite{budai2013robust}, CHASEDB\cite{6224174} for vessel segmentation, four OCTA datasets ROSE-1 \cite{9284503}, ROSE-2 \cite{9284503}, OCTA-6M \cite{li2020image}, OCTA-3M\cite{li2020image} for vessel segmentation, one lesion segmentation dataset iDRiD \cite{data3030025}, and a retinal layer segmentation dataset ARoI \cite{melinvsvcak2021annotated}.


\subsection{Evaluation Metrics}
The metrics to evaluate the segmentation performance are Precision(P), Recall(REC), Dice, Bookmaker Informedness (BM), and Intersection over Union (IoU). The equations are listed as follows:

\textbf{Precision(P):}
\begin{equation}
    P = \frac{TP}{TP + FP} 
\end{equation}


\textbf{Recall(REC):}
\begin{equation}
   REC =  \frac{TP}{TP + FN}
\end{equation}

\textbf{Dice:}
\begin{equation}
   Dice =  \frac{2*TP}{ 2*TP + FN + FP}
\end{equation}

\textbf{Bookmaker Informedness (BM):}
\begin{equation}
    BM = \frac{TP}{TP+FN}+\frac{TN}{TN+ FP}-1 
\end{equation}

\textbf{IoU:}
\begin{equation}
   IoU =  \frac{TP}{TP + FN + FP}
\end{equation}
where $TP,FP,TN$ and $FN$ are "True Positive", "False Positive", "True Negative" and "False Negative", respectively.

\subsection{Implement Details}
We implement the models by Pytorch\cite{NEURIPS2019_9015} framework, and all experiments are run on the machine with one NVIDIA Tesla A100 graphics card. The mini-batch stochastic gradient descent (SGD) with a momentum of 0.9 and a weight decay of 0.0005 is applied to optimize the model. Poly learning rate adjusts strategy \cite{DBLP:journals/corr/LiuRB15} is adopted to set the learning rate dynamically during training, which sets the learning rate according to $lr=init\_lr\times(1-\frac{iter}{max\_iter})^{power}$, and we set $init\_lr=0.05, power=0.9$. The mix-precision training strategy is adopted to save the memory.

\subsection{Experimental Results}
\begin{figure}[htbp]
    \centering
    \includegraphics[width=0.5\textwidth]{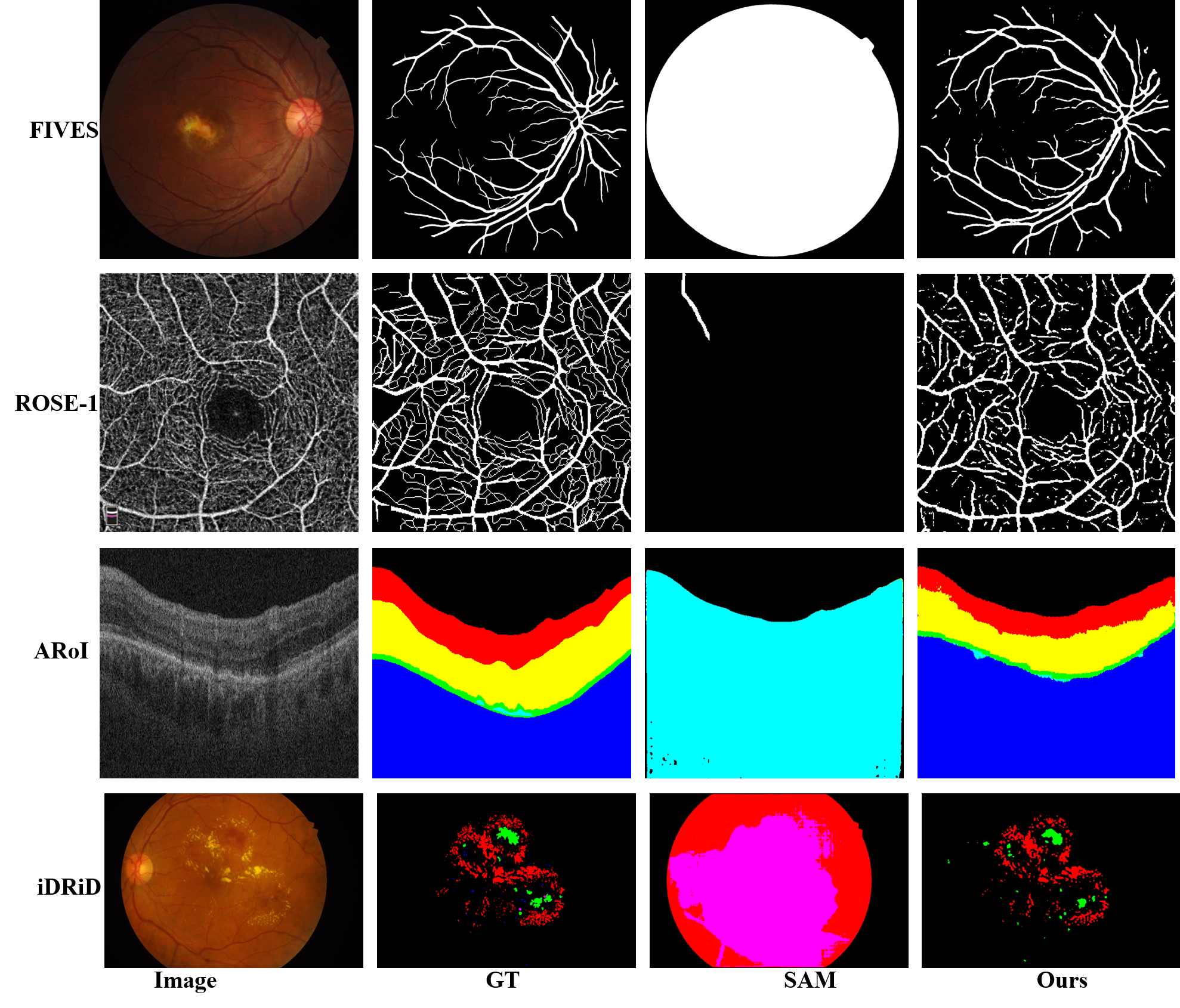}
    \caption{Visualization of the segmentation results on four tasks.}
    \label{fig:cf_vis}
\end{figure}

We first conduct the one-shot learning experiments on three different segmentation tasks: vessel segmentation with two-modal images, lesion segmentation, and OCT layer segmentation. The results are shown in Table \ref{tab:exp_vessel} and Figure \ref{fig:cf_vis}.
For the \textbf{vessel segmentation task}, we use two-modal images including color fundus and OCTA to evaluate the performance. The results are shown in Table \ref{tab:exp_vessel}, from which we can see the SAM with our proposed prompt can provide a comparable performance at the OCTA vessel segmentation task on the ROSE-2 with other supervised models based on the whole dataset including U-Net\cite{10.1007/978-3-319-24574-4_28} and OCTA-Net\cite{9284503} model. 
\begin{table}[htbp]
    \centering
    \caption{Segmentation results based on the one-shot Prompt Learning SAM.$^*$ is the results that trained by one volume of OCT dataset.}
    \begin{tabular}{ccccccc}
     \toprule
    Dataset&Model & Dice & REC & BM & P  \\
     \midrule
        \multirow{2}{*}{FIVES}& SAM & 15.72 & 82.17&15.91&8.69 \\ 
        &SAM(Ours) & 80.90 & 79.53 & 78.22 & 82.31 \\
        \midrule
         \multirow{4}{*}{ROSE-1}&U-Net&66.05&-&-&-\\
        &OCTA-Net&75.76&-&-&-\\
        &SAM&7.50&4.59&0.51&20.59 \\ 
        &SAM (Ours) & 59.01& 58.49& 49.34 & 59.53 \\
        \midrule
        \multirow{3}{*}{ARoI} &SAM&10.42&32.59&25.46&23.02\\
         &SAM(Ours)& 35.68 & 40.55&33.83&40.24 \\
         &SAM(Ours)$^*$&44.59 & 48.82& 45.98 & 43.51 \\ 
     \bottomrule 
           \toprule
      & & MA & HE & EX & SE \\
      \midrule
      \multirow{2}{*}{IDRiD} &SAM&0.14&2.75&3.71&1.42\\
        &SAM(Ours) & 0.00 & 39.54 & 62.96 & 0.00  \\
      \bottomrule
    \end{tabular}
    \label{tab:exp_vessel}
\end{table}

From the qualitative results shown at Figure \ref{fig:cf_vis}, the SAM with our proposed prompt can segment the large vessel in the OCTA correctly, but can not distinct the small or tiny vessels. For the vessel segmentation of color fundus, our SAM with learnable prompt model achieves amazing results on the FIVES dataset. Our model tries to segment the small vessels and distinguish some small vessels but the classification of the small or tiny vessel is still the obstacle to applying the SAM at the vessel segmentation. We will discuss this later. 

We conduct the \textbf{OCT layer segmentation} on the ARoI dataset. As shown in Table \ref{tab:exp_vessel}, the performance of the model raises about 25\% of the dice metric. We visualize the segmentation results in Figure \ref{fig:cf_vis}, which illustrates that the proposed model achieves a good layer results compared to the original model. As OCT images often display as a volume data, we enlarge the size of the data with a volume data to finetune the model. The performance improves further, which may prove that more data for training, the robustness of the model increases.

Finally, we conduct the one-shot \textbf{lesion segmentation} on the iDRiD dataset. As shown in Table \ref{tab:exp_vessel}, although our model fails for the segmentation of microaneurysms (MA) and soft exudates (SE), it works well for the hemorrhages (HE) and hard exudates (EX). The visualization in Figure \ref{fig:cf_vis} also confirms the ability. We analyze that the small medical object is hard to classify by the SAM, because some small objects may be lost during the patch embedding or the edge difference may be not obvious.  

\subsection{Generalization}
To verify the generalization of the model, we conduct a series of zero-shot experiments. Considering the fairness of generalization, we excluded the dataset tasks with excessive diversity when generalizing. Therefore, this paper uses blood vessels on two modalities for generalization experiment verification.

\begin{figure}[htbp]
    \centering
    \includegraphics[width=0.5\textwidth]{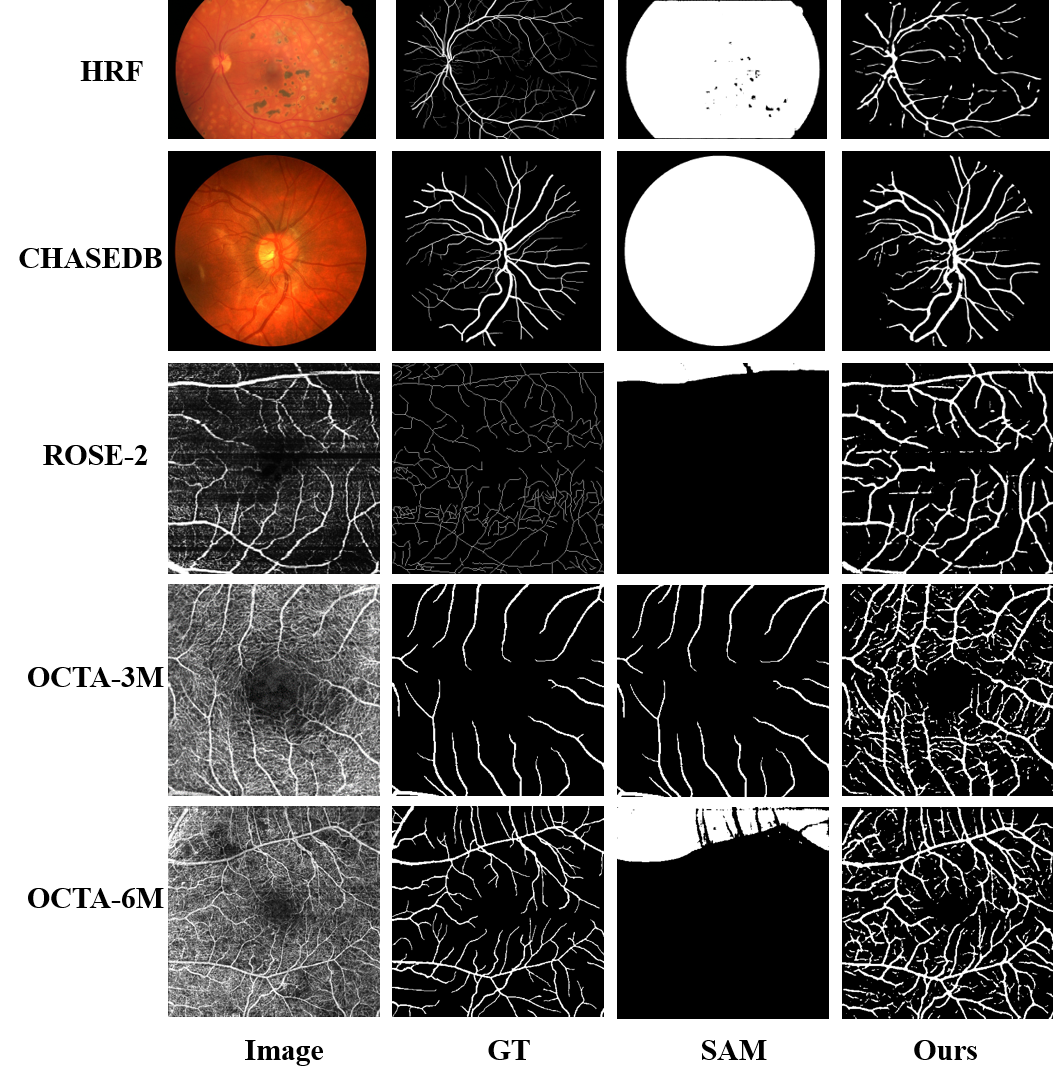}
    \caption{The generalizations results of two-modal images based on five datasets.}
    \label{fig:geralization}
\end{figure}

We conduct the zero-shot experiments for the vessel segmentation on two color fundus datasets. As shown in Table \ref{tab:zero_vessel}, the model trained on the FIVES performs well on the target datasets including HRF and CHASEDB. We find an interesting thing that the model is not disturbed by the lesion in the image on the HRF dataset, as shown in Figure \ref{fig:geralization}. This phenomenon shows the powerful genealization ability of our learnable ophthalmology SAM.
\begin{table}[htbp]
    \centering
    \caption{Zero-shot results for the vessel segmentation.}
    \begin{tabular}{ccccccc}
      \toprule
    Dataset&Model & Dice & REC & BM & P  \\
     \midrule
        
        \multirow{2}{*}{HRF}& SAM & 16.10 & 79.21 & 11.98 & 8.96 \\
        & SAM(Ours)& 69.16 & 76.47& 72.74 & 63.13 \\
        \midrule
        \multirow{2}{*}{CHASEDB}&SAM&8.16&29.74&-14.90&4.73\\
        &SAM(Ours)&76.37&82.98&80.42&70.74\\
        \midrule
        \multirow{2}{*}{ROSE-2}
        &SAM&6.36&5.30&1.14&7.96\\ 
        &SAM(Ours)& 23.33& 41.65 &27.01& 16.20  \\
        \midrule
        \multirow{2}{*}{OCTA-3M}&SAM&8.47&9.52&1.38&7.63\\ 
        &SAM(Ours)&54.10&84.99&75.86&39.68\\
        \midrule
        \multirow{2}{*}{OCTA-6M}&SAM&4.31&2.84&-0.10&8.94\\
        & SAM(Ours) &55.28&77.63&67.14&42.93\\
     \bottomrule
    \end{tabular}
    \label{tab:zero_vessel}
\end{table}

Then we use the ROSE-2, OCTA-3M, and OCTA-6M to do the vessel segmentation on OCTA images, as shown in Table \ref{tab:zero_vessel}. The model trained on ROSE-1 achieves good performance on different datasets captured by the same and different manufactures. The visualization results are shown in Figure \ref{fig:geralization}. OCTA-3M and OCTA-6M are captured by OCTA device from Optovue manufacture, whose images holds more background, while ROSE datasets are by that from Zeiss manufacture, whose background seems clean. The results tell that the model produces good vessel segmentation results on different OCTA datasets without training. Therefore, our learnable ophthalmology SAM performs well based on kinds of segmentation datasets. 

\section{Discussion}
\label{Discussion}
\begin{figure}[htbp]
    \centering
    \includegraphics[width=0.45\textwidth]{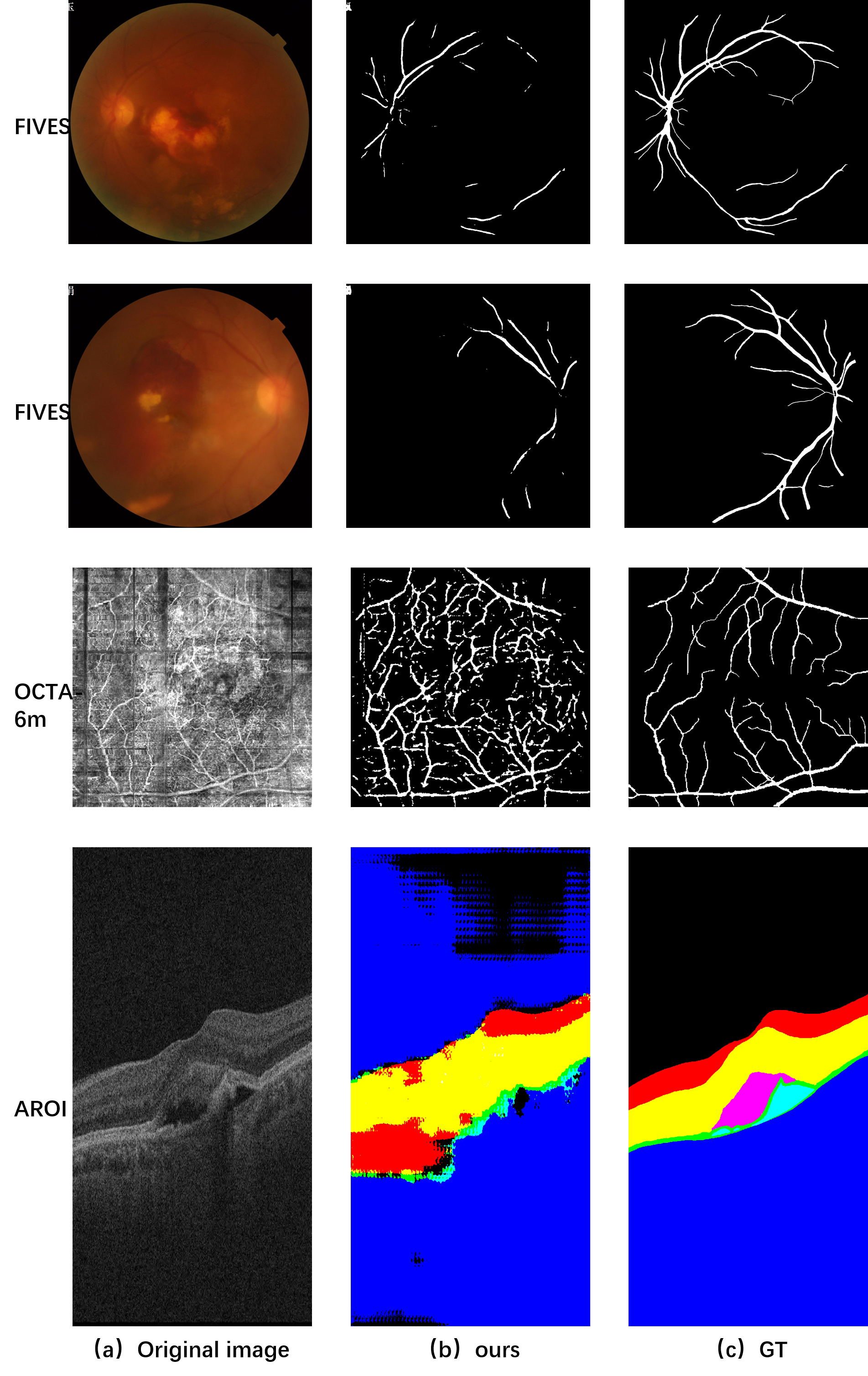}
    \caption{Some failed examples caused by the image quality.}
    \label{fig:blurdis}
\end{figure}

In the experiments, we list the results of the learnable prompt inserted in SAM, which prove the effectiveness of our thought. During experiment, we find that the image quality seriously effects the segmentation results, as shown in Fig. \ref{fig:blurdis}. In FIVES dataset, there are several images with blurred blood vessels, or obscured macular area. Our algorithm only segments parts of the blood vessels. Since OCTA is produced by the combination of variety of B-scan, the image quality is often effected by the eye moving, such as the example in the second row of Fig. \ref{fig:blurdis}. Our algorithm considers some other background as blood vessels. Low image quality of OCT is often caused by the patients' inactive cooperation. Our algorithm fails in such low-quality condition, which still hinders intelligent medical image analysis.

\begin{figure}[htbp]
    \centering
    \includegraphics[width=0.45\textwidth]{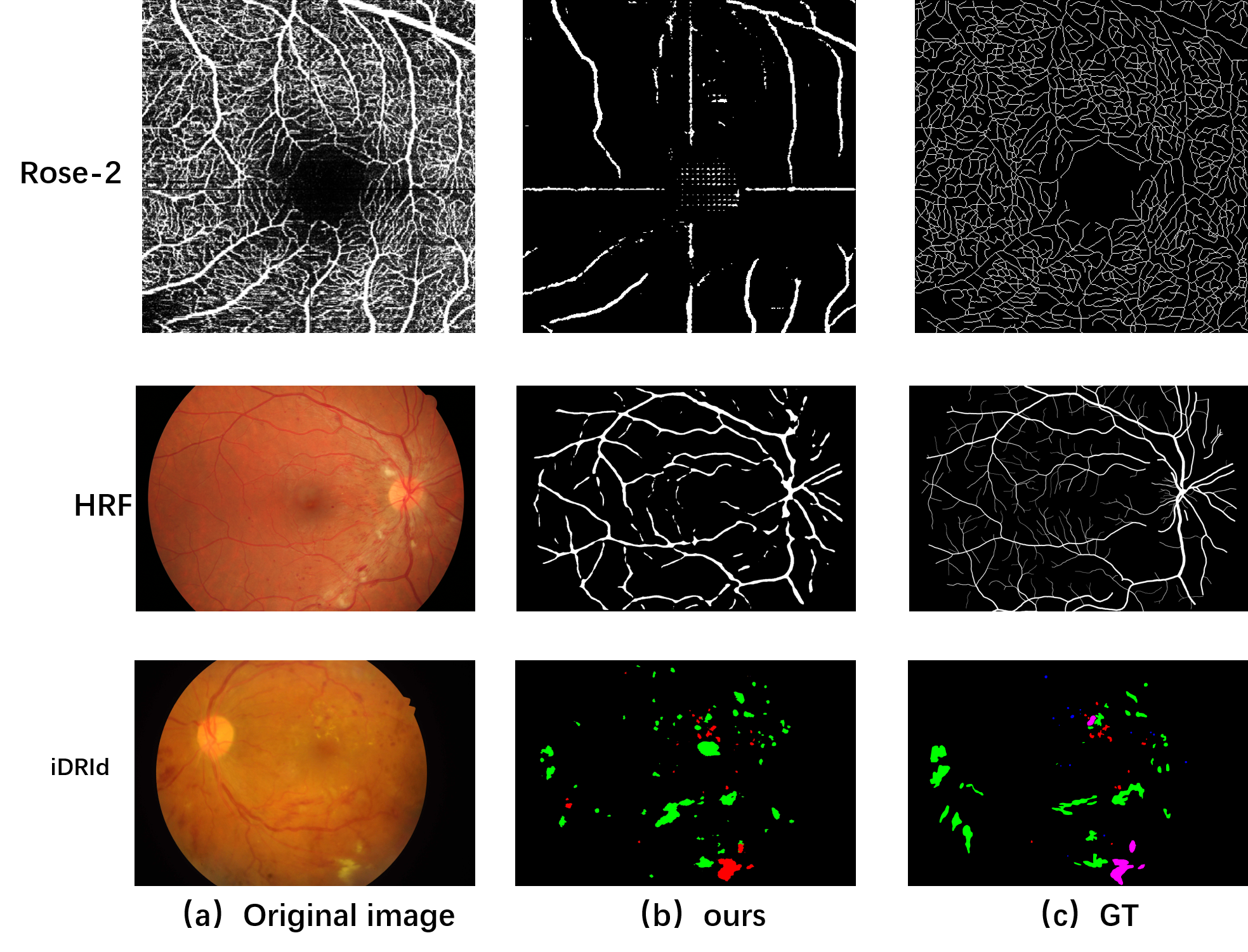}
    \caption{Some failed examples caused by the tiny targets.}
    \label{fig:tinydis}
\end{figure}

There are various of tiny targets in medical images, such as tiny blood vessels or lesions. The segmentation accuracy is not high if the targets are very tiny, as shown in Fig. \ref{fig:tinydis}. In the future, maybe we need to improve the prompt to intensify the tiny target to deal with such condition.

Based on the above experiments, the feature extraction ability of SAM is further proved. The uneven quality of medical images, large modal differences, various analysis targets, data privacy protection, and other issues make it difficult for SAM to be fully applicable in the medical field. Therefore, in order to assist doctors in diagnosis and treatment, it is still necessary to improve the accuracy of target segmentation. We need to consider the speciality of medical scenarios, such as the medical prior knowledge, before designing the model algorithm.

\section{Conclusions}
\label{Conclusions}
In this paper, we proposed a learnable ophthalmology SAM to solve the problem that SAM cannot effectively segment blood vessels or lesions in ophthalmology. The experiments proved the ability, effectiveness and generalization of our proposed prompt layer in three segmentation tasks based on nine publicly available datasets. We also analyzed the failure examples caused by the image quality or targets in the discussion. We also believe that our proposed prompt can be applied in other medical field.

\section{Fund}
This work was supported in part by The National Natural Science Foundation of China (8210072776), Guangdong Provincial Department of Education (2020ZDZX3043), Guangdong Basic and Applied Basic Research Foundation (2021A1515012195), Guangdong Provincial Key Laboratory (2020B121201001), and Shenzhen Stable Support Plan Program (20220815111736001).




\bibliographystyle{elsarticle-num}
\bibliography{prl}







\end{document}